\pdfoutput=1

\documentclass[11pt]{article}

\usepackage[final]{acl}

\usepackage{times}
\usepackage{latexsym}

\usepackage[T1]{fontenc}

\usepackage[utf8]{inputenc}

\usepackage{microtype}

\usepackage{inconsolata}

\usepackage{graphicx}

\usepackage{color}
\usepackage{array}
\usepackage{tabularx}
\usepackage{tabu}
\usepackage{booktabs}
\usepackage{multirow}

\usepackage{graphicx} 
\usepackage{float}
\usepackage{subfigure} 
\usepackage{amssymb}

\usepackage{hyperref}
\usepackage{tablefootnote}
\usepackage{xspace}

\usepackage{float}
\usepackage{amsmath}
\usepackage{bm}
\usepackage{algorithmicx}
\usepackage{algorithm}
\usepackage{algpseudocode}

\usepackage{arydshln}

\usepackage{amssymb}
\usepackage{pifont}
\usepackage{enumitem}
\definecolor{backred}{RGB}{255, 190, 190}
\definecolor{backblue}{RGB}{210, 230, 250}
\usepackage{microtype, tcolorbox}
\usepackage{fontawesome5}

\newcommand{\ours}{\textsc{FinLFQA}\xspace}
\newcommand{\nexample}{1,008\xspace}

\newcommand{\github}{\raisebox{-1.5pt}{\includegraphics[height=1.05em]{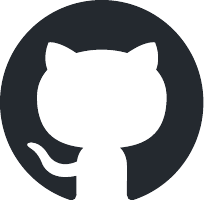}}\xspace}

\title{\ours: Evaluating Attributed Text Generation of LLMs in \\ Financial Long-Form Question Answering}

\author{Yitao Long$^{3}$\thanks{Equal Contributions. Correspondence: Yilun Zhao (\texttt{yilun.zhao@yale.edu}), Chen Zhao (\texttt{cz1285@nyu.edu})}\quad
  Tiansheng Hu$^{1*}$ \quad
  Yilun Zhao$^{2\dagger}$ \quad
  Arman Cohan$^{2}$ \quad
  Chen Zhao$^{1,3\dagger}$ 
  \vspace{5pt}
  \\
  $^1$NYU Shanghai \quad $^2$Yale University \quad $^3$ New York University \vspace{3pt}\\
\github ~~~\url{https://github.com/yitaoLong/FinLFQA}
}

\begin{document}
\maketitle
\begin{abstract}

Large Language Models (LLMs) frequently hallucinate to long-form questions, producing plausible yet factually incorrect answers. 
A common mitigation strategy is to provide attribution to LLM outputs. However, existing benchmarks primarily focus on \emph{simple attribution} that retrieves supporting textual evidence as references. We argue that in real-world scenarios such as financial applications, attribution goes beyond reference retrieval.
We introduce \ours, a benchmark designed to evaluate the ability of LLMs to generate long-form answers to complex financial questions with reliable and nuanced attributions. \ours evaluates three critical aspects of attribution through human annotations: (1) supporting evidence extracted from financial reports, (2) intermediate numerical reasoning steps, and (3) domain-specific financial knowledge that informs the reasoning process.
We further provide an automatic evaluation framework covering both answer quality and attribution quality. Through extensive experiments on eight LLMs across multiple attribution-generation paradigms, we find that 
fine-grained metrics are important to distinguish model capabilities, that 
end-to-end generation achieves comparable performance to post-hoc approaches, and that iterative refinement only helps when guided by external feedback. 


\end{abstract}
\section{Introduction}
\begin{figure*}[!t]
    \centering
    \includegraphics[width = \linewidth]{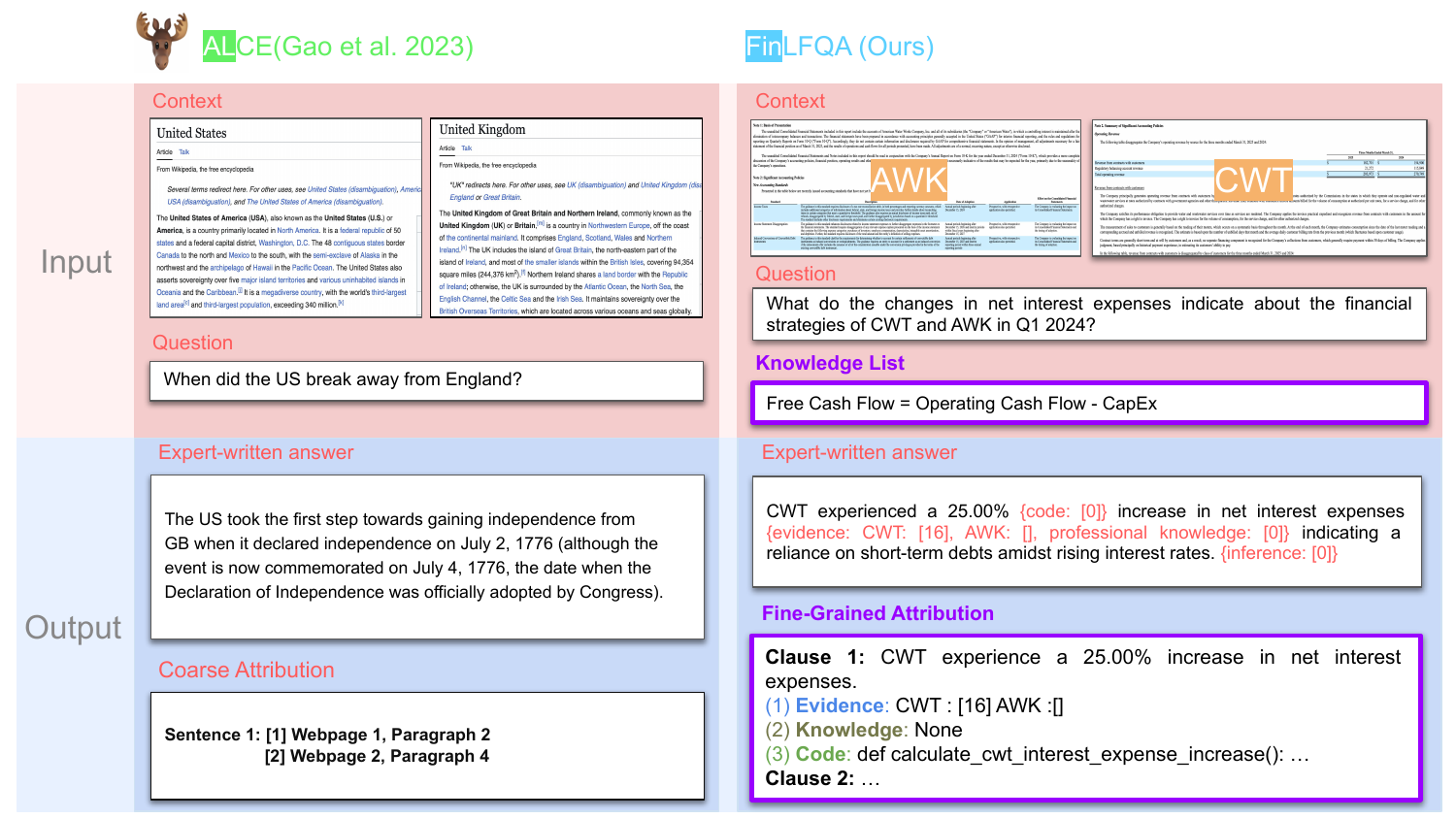}
    \caption{\textbf{(Left)} Compare to previous dataset~\cite{gao-etal-2023-enabling} on long form question answering with annotations, \ours features clause level attribution, generation with knowledge retrieval and multi-faceted attribution.
    \textbf{(Right)} Overview of \ours. The input consists of: (1) context—financial report paragraphs from two companies, (2) a question, and (3) a list of professional knowledge entries that may help in answering about the financial question. The outputs include: (a) an expert-written answer to the question by our annotators, and (b) clause-level attributions, which cover three aspects: \emph{Evidence} (paragraph indices supporting the answer),  \emph{Knowledge} (entries from the provided knowledge list used), and \emph{Code} (a Python snippet used to compute the numerical result when the answer involves calculations). }
    \label{fig:evaluations}
\end{figure*}

Long-form question answering (LFQA) over documents poses a substantial challenge for current large language models (LLMs) because it demands the ability to process and retain information across lengthy contexts, perform multi-step reasoning, and generate factually accurate responses \cite{xu-etal-2023-critical, gao-etal-2023-enabling, zhao2025sciarena}.  %
A key challenge in LFQA is hallucination, where LLMs produce output that is not grounded in the source content, which can severely compromise user trust \cite{ji2023survey}.

To address this challenge, researchers have demonstrated increasing interest in attributed text generation~\cite{gao-etal-2023-enabling, ye-etal-2024-effective}, which aims to improve the trustworthiness of generated content by providing supporting evidence for model outputs. This attribution enables users to verify claims and assess the reliability of responses. Most current approaches focus on tasks such as fact-checking~\cite{kamoi-etal-2023-wice} and summarization~\cite{huang2024learningfinegrainedgroundedcitations}, where evidence attribution is comparatively straightforward.

However, existing benchmarks ~\cite{kamoi-etal-2023-wice, gao-etal-2023-enabling} primarily evaluate attribution through a single lens: evidence retrieval. While this is a necessary component, attribution in critical domains like finance extends well beyond locating supporting passages. As shown in Figure~\ref{fig:evaluations}, two additional forms of attribution are important in financial domain. First, financial texts are rich in numerical data, requiring models not only to identify evidence but also to execute precise numerical reasoning to derive trustworthy conclusions~\cite{zhao-etal-2024-docmath}. Second, reliable answers demand integration of specialized financial knowledge, where models must grasp intricate domain concepts and relationships~\cite{gan2024mmefinancemultimodalfinancebenchmark}.


To address these gaps, we introduce \ours, a comprehensive benchmark for evaluating long-form QA and attributed generation by LLMs. 
\ours consists of \nexample expert-annotated instances.
As illustrated in Figure \ref{fig:evaluations}, it is designed to test LLMs' ability to apply financial knowledge, perform analytical reasoning, and generate detailed answers grounded in the financial reports of two relevant companies.
In contrast to prior work that focuses primarily on evidence attribution, \ours evaluates three distinct forms of attribution in generated responses: \emph{supporting evidence}, \emph{intermediate reasoning steps}, and \emph{domain-specific knowledge}. These attribution types are essential in the financial domain and broadly relevant to other high-stakes applications.

To assess model performance, we design an automatic evaluation system that extends beyond surface-level metrics such as ROUGE or BERTScore. Our framework introduces fine-grained dimensions that jointly evaluate factual accuracy, numerical correctness, and attribution quality across evidence, reasoning, and domain knowledge. Using this system, we benchmark a range of LLMs under multiple attribution paradigms. Results show that proprietary models achieve the strongest overall performance, while leading open-source models are increasingly competitive. End-to-end generation performs on par with post-hoc attribution, whereas iterative refinement yields limited gains without external feedback, though it improves code execution reliability.


Our contributions are summarized as follows:

\begin{itemize}[leftmargin=*]
    \item We propose a comprehensive benchmark, \ours, specifically designed for evaluating long-form question answering and attribution generation in the financial domain. 
    \item We design an automated evaluation system with fine-grained metrics to better capture financial reasoning and precision.
    \item We implement three attribution pipelines—\emph{post-hoc}, \emph{end-to-end}, and \emph{iterative refinement}—and show that end-to-end matches post-hoc in performance, while iterative refinement only improves with external feedback.
\end{itemize}

\section{Related Work}
\subsection{Long-form Question Answering}
Hallucination is a significant issue for generative LLMs \cite{xiao-wang-2021-hallucination, shuster-etal-2021-retrieval-augmentation}. Many studies have explored augmenting LMs with externally retrieved information to mitigate this problem \cite{pmlr-v162-borgeaud22a, JMLR:v24:23-0037}. This issue has also attracted growing interest in attributed LLMs \cite{hu2024benchmarkinglargelanguagemodels, 10516637}, which aim to enhance the verifiability of information by generating responses that attribute content to reliable sources.
However, most existing work has focused on tasks such as question answering or text summarization in information-seeking contexts \cite{bohnet2023attributedquestionansweringevaluation, xu2024aliiceevaluatingpositionalfinegrained}. These approaches often overlook more complex question-answering scenarios that involve numerical reasoning and knowledge-intensive tasks in real-world settings. Therefore, in our work, we will incorporate knowledge into the prompts, requiring models to generate programs that show calculation steps and demonstrate knowledge usage, accompanied by appropriate citations.

\subsection{Attributed Text Generation}
Attribution has gained significant attention for enhancing the interpretability, verifiability, and safety of LLMs \cite{li2023surveylargelanguagemodels, cohenwang2024contextciteattributingmodelgeneration}. Two main approaches have emerged in this field. The first approach utilizes prompt-based or in-context learning, where LLMs are instructed to generate responses with citations given the question and retrieved paragraphs \cite{kamalloo2023hagridhumanllmcollaborativedataset, gao-etal-2023-enabling}. The second approach employs post-hoc methods, which attribute existing responses using an auxiliary language model to identify relevant sources \cite{gao-etal-2023-rarr,huang-etal-2024-advancing}. \citet{ye-etal-2024-effective} demonstrated that fine-tuning an LLM to generate citations and iteratively refine responses yields more accurate results than both prompting-based and post-hoc methods.
However, existing work on attribution for language models has primarily focused on citation generation. In the financial domain, citation attribution alone is insufficient to mitigate hallucination and generate robust results, as financial tasks often require numerical reasoning and domain-specific knowledge. Therefore, \ours includes not only citation attribution but also numerical reasoning processes and financial domain expertise.
\begin{figure*}[!t]
    \centering
    \includegraphics[width = \linewidth]{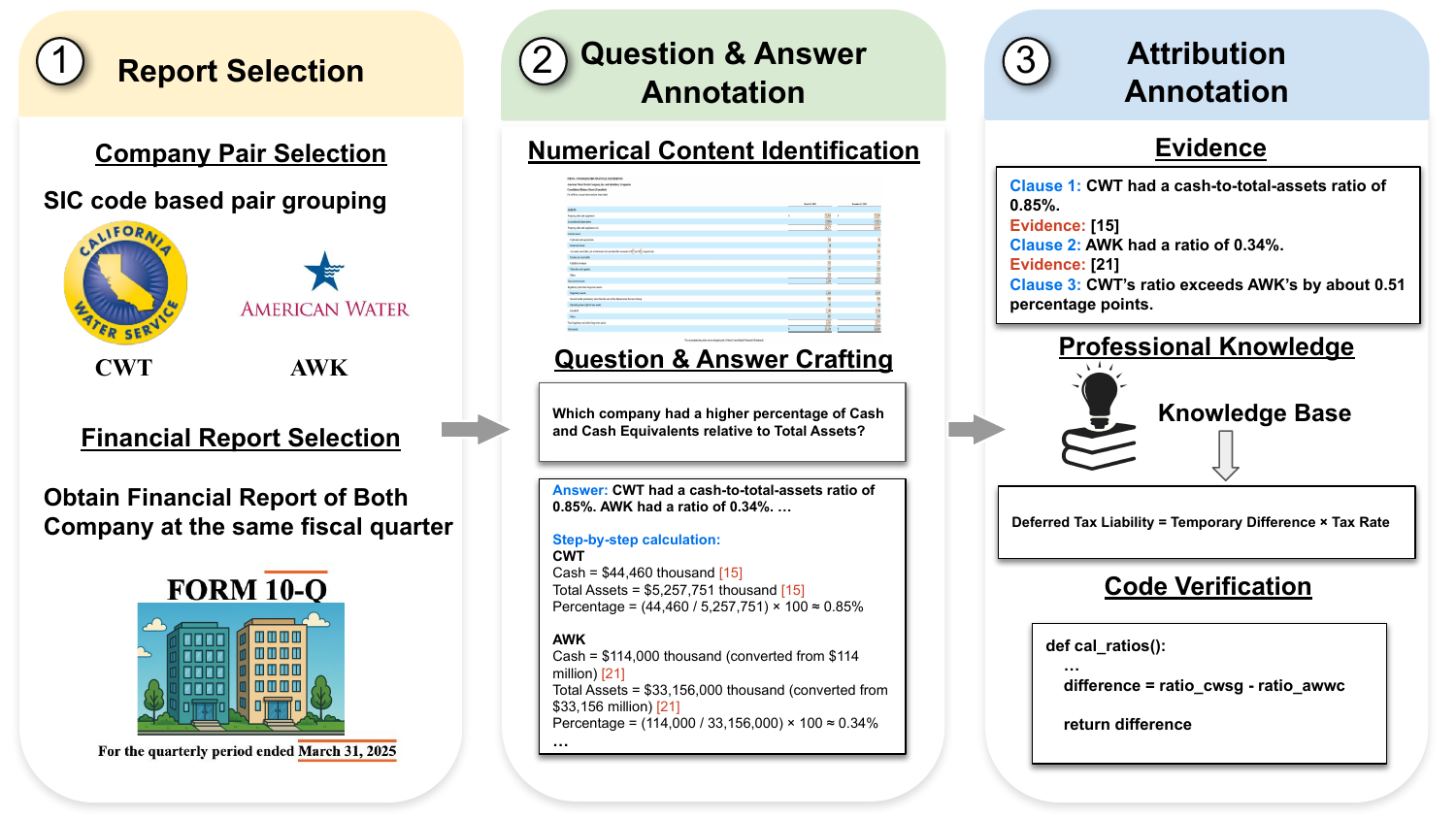}
    \caption{
        Overview of 
        the three-stage process for \ours construction.
        \textbf{(1) Report Selection:} We select company pairs based on their SIC codes and obtain their financial reports for the same fiscal quarter.
        \textbf{(2) Question \& Answer Annotation:} We then identify key numerical content from both financial reports. Given those information, the annotators craft calculation-based questions requiring cross-company and multi-source reasoning, and providing detailed, step-by-step answers citing relevant paragraphs.
        \textbf{(3) Attribution Annotation:} Finance experts verify and split answers into evidence-backed clauses, annotate relevant professional financial concepts from a knowledge base, and translate verified calculations into structured Python functions for reproducibility and validation.
    }
    \label{fig:annotation}
\end{figure*}

\section{\ours Benchmark}
In this section, we first formulate the task, then present the statistical analysis of the dataset and describe our data annotation process. Table \ref{profile} in the Appendix presents the profiles of the eleven annotators involved.

\subsection{Task Formulation}
We define the task of \ours as follows: Given portions of financial documents from two companies\footnote{Our main goal is to evaluate models’ ability to generate attributable financial report analysis. Starting with two companies provides a manageable and meaningful testbed. As models improve, this can extend to include more companies.} comprising multiple text paragraphs in set $\mathcal{D}$, where each document contains both textual and tabular data, and a query $q$, the model should generate a response $\mathcal{R}$ based on the provided context. 
The response consists of $n$ statements ${s_1, s_2, \ldots, s_n}$, where together these statements form the answer to the query. Each statement $s_i$ consists of:
\noindent (a) A clause $t_i$ that contributes to answering the query;
\noindent (b) Three types of attributions: 
\noindent (1) A list of paragraph indices $\mathcal{C}_i = {i_1, i_2, \ldots, i_k}$ where $i_j \in {1, 2, \ldots, |\mathcal{D}|}$, indicating the relevant source paragraphs supporting the statement.
\noindent (2) An intermediate reasoning process $\mathcal{P}_i$ expressed as a Python program, if the statement involves numerical reasoning.
\noindent (3) A list of professional knowledge indices $\mathcal{K}_i = {k_1, k_2, \ldots, k_m}$ where $k_j \in {1, 2, \ldots, |\mathcal{K}|}$, referencing entries in the knowledge base $\mathcal{K}$. Thus, each statement can be represented as $s_i = (t_i, \mathcal{C}_i, \mathcal{P}_i, \mathcal{K}_i)$, and the optimal response can be formulated as:
\begin{equation}
\mathcal{R}^* = \underset{\mathcal{R}}{\operatorname{argmax}} \ P(\mathcal{R}|q, \mathcal{D}, \mathcal{K})
\end{equation}
where $P(\mathcal{R}|q, \mathcal{D}, \mathcal{K})$ represents the probability of generating response $\mathcal{R}$ given the query $q$, document set $\mathcal{D}$, and knowledge base $\mathcal{K}$.

\subsection{Data Annotation}
\label{data_annotation}
Figure \ref{fig:annotation} presents the overview of \ours construction process.

\paragraph{Report Selection.} Following previous work \cite{zhao-etal-2024-docmath, zhao-etal-2024-findver}, we use quarterly reports (i.e., Form 10-Q) of companies as our source documents, which are publicly available in the U.S. Securities and Exchange Commission's open-source database\footnote{https://www.sec.gov/edgar/search/}. We selected two companies based on their Standard Industrial Classification (SIC) Code\footnote{https://www.sec.gov/search-filings/standard-industrial-classification-sic-code-list}, randomly choosing companies within the same code to ensure industry comparability. For these two companies, we chose financial reports from the same fiscal quarter to maintain temporal consistency.


\paragraph{Data Annotation.} Annotators were given financial reports from two companies and instructed to identify sections with numerical values and overlapping content. Based on these, they crafted questions requiring calculations, preferably involving evidence across multiple paragraphs and tables. Priority was placed on cross-company comparisons, integration of table and text data, temporal reasoning (e.g., year-over-year changes), and derived metric computations (e.g., margins, ratios, multi-year averages). We also emphasized diversity across financial topics (e.g., cost structure, income streams, debt levels, investment allocations) to ensure broad coverage of financial reasoning skills. Bonus compensation was provided for complex mathematical reasoning beyond basic arithmetic. After writing each question, annotators structured answers into atomically numbered clauses, each citing supporting paragraphs or prior inferences, with detailed documentation of all computational steps.

\paragraph{Attribution Annotation.}
Finance-expert annotators verify the initial annotations, ensuring that answers are clear, well-supported, and entirely derivable from the provided text.
Upon passing quality checks, annotators proceed with evidence attribution annotation. For the professional knowledge component, we randomly sample one thousand financial concepts from our Wikipedia-based knowledge base with the size of twenty for each question. Annotators then indicate whether any of these concepts inform the reasoning in each clause. For the calculation process annotation, financial experts first verify the mathematical equations, which are then passed to finance-expert annotators who convert the verified calculations into Python functions. These functions are structured to include variable assignments, calculation steps, and return values.

\subsection{Data Statistics}
\begin{table}[t!]
\centering
\small
\renewcommand{\arraystretch}{1.1} 
\setlength{\tabcolsep}{4pt} 
\begin{tabular}{lr}
\toprule
\textbf{Property} \texttt{(Median/Avg)} & \textbf{Value} \\ 
\midrule
Question Length  & 16.0 / 16.3 \\
Answer Length  & 51.0 / 52.4 \\
\# Clauses & 3.0 / 3.4 \\
Report Length & 2144.0 / 2221.6 \\
\# Paragraphs & 40.0 / 40.5 \\
\# Companies & 89 \\
\hdashline
\# Evidence & 2.0 / 2.9 \\
\# Code & 1.0 / 1.1 \\
\# Professional Knowledge & 1.0 / 1.2 \\
\hdashline
Development Set Size & 302 \\
Test Set Size & 706 \\
\bottomrule
\end{tabular}
\caption{Data statistics in \ours dataset.}
\label{dataset_statistics}
\end{table}

Table \ref{dataset_statistics} summarizes the statistics of \ours, which comprises 1008 examples. The dataset is randomly split into a development set (302 examples, 30\%) and a test set (706 examples, 70\%). To avoid data contamination, test set answers remain private. Instead, an online evaluation platform enables researchers to assess models and join a leaderboard. Evaluation is performed in a zero-shot setting.

\section{Automated Evaluation System}

Evaluating financial long-form question answering is challenging because it requires assessing textual accuracy, numerical correctness, and domain knowledge. 
To address these challenges, in addition to traditional overlap-based metrics (i.e., ROUGE and BERTScore), we propose fine-grained metrics that jointly capture these dimensions.

\subsection{LLM-as-Judges for Answer Evaluation}


This approach \cite{10.5555/3666122.3668142} leverages LLMs as automated evaluators, offering a scalable alternative to costly human evaluation while maintaining assessment quality. Unlike ROUGE and BERTScore, which rely solely on surface-level or embedding-based similarity measurements, LLM-based evaluation can assess nuanced aspects of answer quality. We evaluate responses by providing GPT-4o with the full context of financial reports, question, generated answer, and ground truth. The model assesses three key criteria: (1) \textbf{Accuracy}: whether the answer correctly addresses the question and aligns with the ground truth; (2)  \textbf{Numerical Correctness}: whether all numerical calculations and values are precise and accurate; and (3)  \textbf{Evidence Entailment}: whether all claims are properly substantiated by information from the financial reports. Each criterion is scored on a scale of 1 to 5, with the final score being the sum of these three components, ranging from 3 to 15.

\paragraph{Evaluating System Reliability.}
To assess the reliability of our automated evaluation system, we conducted a human expert study on 50 randomly selected examples. Each response was independently rated by two annotators with finance backgrounds, using the same evaluation criteria as our LLM-as-judge framework. We computed Pearson correlations between human scores and GPT-4o scores for each dimension, obtaining 0.853 for overall score, 0.812 for answer accuracy, 0.807 for numerical correctness, and 0.630 for evidence entailment. These results indicate strong alignment between human judgment and the LLM-as-judge system, supporting the reliability of our evaluation pipeline.


\subsection{Numerical Reasoning Assessment} 
In financial analysis, numerical precision is critical, even when the semantic content of the answer is preserved. Small numerical discrepancies can lead to substantial misinterpretations, making traditional semantic similarity metrics insufficient. To address this, we evaluate numerical accuracy by extracting all numerical values from both the ground truth and model-generated answers.
We compute \textbf{precision}, \textbf{recall}, and \textbf{F1 score} as follows: let $G$ denote the ground truth set (e.g., correct numerical values) and $M$ denote the model-predicted set. Precision is defined as $|M \cap G| / |M|$, representing the fraction of predicted items that are correct. Recall is defined as $|M \cap G| / |G|$, representing the fraction of ground truth items that are correctly predicted. F1 score is the harmonic mean of precision and recall, calculated as $(2 \cdot \text{Precision} \cdot \text{Recall}) / (\text{Precision} + \text{Recall})$. 

To account for real-world variations in numerical representation, we implement flexible matching strategies. First, for rounding tolerance, a predicted number is considered correct if it matches the ground truth within a relative tolerance of 0.01 (e.g., 3.965 is matched to 3.97). Second, for scale normalization, we standardize values across commonly used financial scales, treating numbers like 3 million, 3,000 thousands, and 3,000,000 as equivalent. Additionally, normalization is applied using scale factors from the set \{1, 100, 1000, 1000000, 0.01, 0.001, 0.000001\}.

\subsection{Fine-Grained Attribution Evaluation}
We also design a set of fine-grained metrics to assess the key dimensions of the generated answers.

\paragraph{Evidence.}
We evaluate the quality of evidence attribution by measuring how well the model identifies and cites relevant supporting evidence from the financial reports. For each generated answer, we compute \textbf{precision}, \textbf{recall}, and \textbf{F1 score}. 

\paragraph{Code.}
We evaluate the model's ability to generate executable code for numerical calculations by measuring the \textbf{execution success rate}. This metric calculates the percentage of generated code snippets that successfully execute. This evaluation ensures that the model not only provides correct answers but also generates valid, executable code to support its calculations.

\paragraph{Professional Knowledge.}
We assess the model's use of professional financial knowledge through a \textbf{recall}-based evaluation. This measures how well the model identify knowledge that contributes reasoning of statements.

\section{Experiments}
This section describes the experimental setup, models evaluated, main results, and error analysis.

\subsection{Evaluated LLM-based Systems}

In \ours, we require LLMs to not only generate answers to queries but also provide three distinct attributions for each response. To achieve this, we propose three approaches: post-hoc generation, end-to-end generation, and iterative refinement.

\paragraph{Post-hoc Generation.}
\label{post_hoc}
The post-hoc generation process consists of two stages. In the first stage, given the financial reports and question as input, the model generates an answer. In the second stage, we generate attributions post-hoc: given the reports, question, and the previously generated answer, the model produces three attributions (evidence, numerical reasoning and professional knowledge integration) for each clause in the response. The prompts for answer generation and attribution generation are shown in Figure \ref{post_hoc_ans} and Figure \ref{post_hoc_attr}.

\paragraph{End-to-end Generation.}
\label{end2end}
In end-to-end generation, LLMs produce both the response clauses and their attributions simultaneously in a single pass. Given a financial report and a question as input, the model first generates an answer to the question, followed by the three attribution types for each clause in the response. The prompt is shown in Figure~\ref{e2e_generation}.

\paragraph{Iterative Refinement.} 
\label{iterative}
Iterative generation introduces a multi-step refinement process to enhance response accuracy and reliability. The LLM first generates an initial response. Next, the same LLM extracts Python code blocks, executes them in a Python environment to obtain computed values, and incorporates these values into the response. Then, given the financial reports, question, initial response, and execution results, the same LLM provides feedback based on four criteria: (1) completeness—whether the response fully addresses the question; (2) evidential support—ensuring every claim is backed by the financial reports to mitigate hallucination and enhance robustness; (3) numerical consistency—verifying that the extracted program output aligns with the values stated in the response, and if execution fails, providing debugging guidance; and (4) professional knowledge integration—assessing whether the identified domain knowledge contributes to reasoning in the clause. 
Based on this evaluation, the model generates structured feedback to guide the next iteration. This iterative refinement continues until either no further improvements are identified or the process reaches a predefined maximum iteration threshold.
The prompt is shown in Figure \ref{e2e_feedback}.

\begin{table*}[!t]
\centering
\resizebox{0.95\textwidth}{!}{%
\addtolength{\tabcolsep}{0.1em}
\begin{tabular}{lccccccccccc}
\toprule
 \multirow{3}{*}{\textbf{Model}} & \multicolumn{5}{c}{\textbf{Attribution Performance}} & \multicolumn{6}{c}{\textbf{Answer Performance}} \\
\cmidrule(lr){2-6} \cmidrule(lr){7-12} 
 
 & \multicolumn{3}{c}{\textbf{Evidence}} & \textbf{\% Code} & \textbf{Knowl.} & \multirow{2}{*}{\textbf{R-L}} & \multirow{2}{*}{\textbf{BS}} & \textbf{LLM-} & \multicolumn{3}{c}{\textbf{Numerical}}\\
\cmidrule(lr){2-4} \cmidrule(lr){10-12} 
& \textbf{Prec.} & \textbf{Recall} & \textbf{F1} & \textbf{Exec. Rate} & \textbf{Recall} & & & \textbf{as-Judge} & \textbf{Prec.} & \textbf{Recall} & \textbf{F1}\\
\midrule
\multicolumn{12}{c}{\emph{\textbf{Post-hoc}}} \\\noalign{\vskip 0.5ex}
GPT-4o & 49.0 & \textbf{78.1} & 55.5 & 9.3 & 19.3 & 23.1 & 87.5 & 13.6 & \textbf{37.9} & \textbf{58.0} & \textbf{42.3} \\
Qwen2.5-72B & 39.5 & 71.1 & 45.7 & 10.3 & 19.1 & 22.6 & 87.9 & 13.0 & 33.7 & 52.8 & 37.8 \\
Llama-3.3-70B & 44.9 & 71.3 & 50.4 & 16.6 & 16.1 & 19.7 & 87.5 & 13.1 & 35.9 & 51.8 & 39.1 \\
Llama-3.2-1B & 1.7 & 3.0 & 1.9 & 1.5 & 0.5 & 15.3 & 84.6 & 5.6 & 12.3 & 14.5 & 11.7 \\
Llama-3.2-3B & 10.5 & 19.2 & 12.2 & 14.3 & 8.2 & 19.1 & 86.6 & 8.9 & 24.2 & 32.6 & 25.3 \\
Mistral-Small-24B & 49.1 & 69.9 & 53.0 & 4.5 & 22.2 & 25.0 & 88.2 & 12.6 & 36.3 & 48.7 & 38.5 \\
Mistral-8x22B & 36.2 & 58.5 & 40.9 & 11.1 & 16.7 & 21.7 & 87.3 & 12.1 & 33.7 & 42.6 & 34.1 \\
phi-4 & 42.1 & 66.8 & 47.3 & 8.1 & 18.3 & 20.1 & 87.0 & 13.0 & 32.5 & 48.8 & 35.6 \\
\midrule
\multicolumn{12}{c}{\emph{\textbf{End-to-end Generation}}} \\\noalign{\vskip 0.5ex}
GPT-4o & 46.9 & 75.0 & 53.3 & 26.3 & 17.3 & 19.3 & 87.0 & \textbf{13.7} & 35.2 & 54.9 & 39.4 \\
Qwen2.5-72B & 48.7 & 68.6 & 50.4 & 15.4 & 17.1 & 21.7 & 86.9 & 12.1 & 33.4 & 45.3 & 36.1 \\
Llama-3.3-70B & \textbf{53.7} & 68.3 & \textbf{56.1} & 20.1 & 17.1 & 21.1 & 87.0 & 12.0 & 35.8 & 45.0 & 36.5 \\
Llama-3.2-1B & 0.6 & 0.7 & 0.6 & 8.5 & 0.0 & 13.9 & 81.1 & 4.5 & 6.8 & 8.5 & 6.6 \\
Llama-3.2-3B & 13.3 & 22.8 & 15.1 & 22.5 & 6.2 & 18.5 & 85.8 & 7.5 & 19.3 & 27.0 & 19.9 \\
Mistral-Small-24B & 52.9 & 69.3 & 55.8 & 12.9 & \textbf{22.4} & 23.3 & 87.4 & 12.4 & 33.8 & 47.5 & 36.5 \\
Mistral-8x22B & 40.1 & 57.8 & 41.2 & 18.4 & 16.5 & 21.4 & 86.6 & 12.0 & 33.6 & 43.7 & 37.1 \\
phi-4 & 40.7 & 59.2 & 44.5 & 14.1 & 16.7 & 21.2 & 86.6 & 12.5 & 37.4 & 45.9 & 37.3 \\
\midrule
\multicolumn{12}{c}{\emph{\textbf{Iterative Refinement}}} \\\noalign{\vskip 0.5ex}
GPT-4o & 46.7 & 75.0 & 53.1 & \textbf{29.8} & 17.1 & 20.0 & 86.8 & 13.5 & 34.2 & 54.7 & 38.9 \\
Qwen2.5-72B & 46.6 & 68.0 & 52.0 & 22.7 & 17.2 & 22.2 & 86.6 & 12.0 & 35.8 & 46.0 & 36.5 \\
Llama-3.3-70B & 50.8 & 67.9 & 54.1 & 27.3 & 17.3 & 22.0 & 86.7 & 12.1 & 35.4 & 46.8 & 36.6 \\
Llama-3.2-1B & 0.5 & 0.7 & 0.5 & 8.9 & 0.0 & 13.0 & 80.3 & 3.8 & 2.8 & 3.0 & 2.3 \\
Llama-3.2-3B & 9.1 & 13.3 & 10.0 & 22.5 & 5.0 & 20.6 & 86.8 & 6.7 & 19.1 & 22.9 & 18.5 \\
Mistral-Small-24B & 53.0 & 68.5 & 55.7 & 18.1 & 18.2 & 25.0 & 88.0 & 12.2 & 33.8 & 46.7 & 36.5 \\
Mistral-8x22B & 43.2 & 57.3 & 46.1 & 18.7 & 16.3 & 25.4 & 88.0 & 12.1 & 33.9 & 47.1 & 37.1 \\
phi-4 & 43.8 & 60.9 & 47.2 & 16.6 & 16.0 & 25.4 & 87.8 & 12.5 & 37.0 & 46.1 & 37.4 \\
\bottomrule
\end{tabular}
}
\caption{Results of \emph{development} set. R-L denotes ROUGE-L, BS denotes BERTScore. The LLM-as-a-judge evaluation uses a 15-point scale, consisting of three main criteria: accuracy, numerical correctness, and supporting evidence. Each criterion is scored from 1 to 5 points. The detailed breakdown of results for each criterion is shown in the Appendix Table \ref{llm_development}.}
\label{development_result}
\end{table*}
\subsection{Implementation Details}
We evaluate eight LLMs on \ours, including GPT-4o, Qwen2.5-72B \cite{qwen2025qwen25technicalreport}, Llama-3.3-70B \cite{grattafiori2024llama3herdmodels}, Llama-3.2-3B, Llama-3.2-1B, Mistral-Small-24B \cite{jiang2023mistral7b}, Mistral-8x22B \cite{jiang2024mixtralexperts}, and Phi-4 \cite{abdin2024phi4technicalreport}. The exact versions and specifications of these models are provided in Table~\ref{tab:model_specification} in the Appendix.

We conducted experiments on open-source LLMs using the vLLM framework \cite{kwon2023efficient}. All experiments were performed under a zero-shot setting, with a temperature of 1.0 and a maximum output length of 2048 tokens.
Following previous work \cite{zhao-etal-2024-docmath, zhao-etal-2024-findver}, we serialize tabular data by using a vertical bar (|) to separate columns and a newline to separate rows.

\subsection{Main Results}
Tables \ref{development_result} and \ref{test_result} present the performance of LLMs on the development and test sets, respectively. Additionally, Tables \ref{llm_development} and \ref{llm_test} provide a detailed breakdown of the LLM-as-a-judge evaluation, including accuracy, numerical correctness, and the quality of evidence entailment. We summarize the main
takeaways from the experiments below.  

\paragraph{Comparing Model Performance.}  
Our results show that GPT-4o achieves the highest score in LLM-as-a-judge evaluation (13.7). It also excels in numerical accuracy, a critical aspect in financial applications, producing responses with more correct numerical values compared to other models. Its precision, recall, and F1 scores in numerical matching are 37.9, 58.0, and 42.3, respectively. In code generation, GPT-4o also leads, with a code execution success rate of 29.8\% in the iterative refinement setting. 
However, strong performance is also observed among open-source models. Qwen2.5-72B shows competitive results across most dimensions, reaching an LLM-as-a-judge score of 13.0 and achieving balanced precision and recall in numerical accuracy (33.7 and 52.8). Llama-3.3-70B consistently delivers solid performance, ranking among the top open-source models in both evidence attribution (e.g., F1 of 56.1 in end-to-end generation) and numerical reasoning (F1 of 39.1). Notably, Mistral-Small-24B performs best in professional knowledge recall (22.4).
Taken together, while GPT-4o leads in overall accuracy and numerical reliability, several open-source models are closing the gap. In particular, Llama-3.3-70B, Qwen2.5-72B, and Mistral-Small-24B are competitive in attribution and reasoning tasks. These results suggest that open-source systems are becoming increasingly viable alternatives given their accessibility, lower cost, and faster inference speed.

\paragraph{Fine-grained evaluation metrics provide better comparison.}  
All models receive low ROUGE scores because it relies on exact match for evaluation, which is less effective for open-ended tasks where multiple valid responses exist. BERTScore, despite capturing semantic similarity, also fails to differentiate model performance meaningfully. In financial applications, numerical values and factual accuracy are crucial, and even minor differences can alter the factual correctness of a response. Since all models achieve similar BERTScores (approximately 88), this highlights its limitations in detecting factual inconsistencies. Therefore, fine-grained evaluation metrics are essential for financial long-form question answering, and we provide results across multiple dimensions.

\paragraph{Post-hoc and end-to-end generation show no significant difference.}  
Our results indicate that post-hoc and end-to-end generation perform similarly, suggesting that current frontier models can handle complex tasks effectively. In the post-hoc approach, models first generate an answer and then attribute supporting evidence, whereas in the end-to-end approach, both are generated simultaneously. The latter does not degrade performance and offers two advantages: (1) generating everything at once reduces computational cost and latency, and (2) it improves code generation consistency. In post-hoc generation, numerical reasoning steps are inferred from a pre-generated answer, leading to inconsistencies and lower execution success rates. In contrast, end-to-end generation ensures that answers and reasoning steps are aligned, making outputs more robust and verifiable.  

\paragraph{Iterative generation via self-feedback does not improve performance.}  
We find no significant improvement on iterative refinement setting. Since this task primarily relies on reasoning abilities, our results align with prior work \cite{huang2024largelanguagemodelsselfcorrect} showing that self-feedback without external signals does not enhance performance. The only observed improvement is in code execution success rate, which can be attributed to feedback from execution results. When errors occur, they provide corrective signals to the model, explaining why execution success improves in the iterative setting. We provide further analysis on \S \ref{iterative_refinement}.

\subsection{Analysis of Iterative Refinement}
\label{iterative_refinement}

Table \ref{tab:iterative_refinement} shows the results of iterative refinement under different settings and model configurations.

\paragraph{Self-feedback alone is insufficient.}
Iterative refinement that relies solely on a model’s own feedback, without any external guidance, does not lead to meaningful performance gains. In fact, we observe consistent degradation across all tested models when comparing the self-feedback setting to the ened-to-end generation. This suggests that models struggle to generate useful self-supervised signals for improving their own outputs, likely due to a lack of external corrective information.

\paragraph{External feedback requires sufficient model capacity.}
While external feedback can enable performance gains, the base model must have enough capacity to utilize it effectively. For instance, Llama-3.2-3B shows improvement when receiving feedback from the stronger Llama-3.3-70B model, whereas the smaller Llama-3.2-1B fails to benefit. This highlights that incorporating external feedback is not universally beneficial—it requires sufficient architectural capacity and representational power in the receiving model to interpret and apply the guidance.

\paragraph{Domain-specific feedback improves outcomes.}
Feedback effectiveness is also influenced by domain alignment. Llama-3.1-8B significantly improves when guided by Fino1-8B \cite{qian2025fino1transferabilityreasoningenhanced}, a model of equal size and architecture but fine-tuned on financial data using reinforcement learning. This result underscores the importance of domain-specific expertise in feedback generators. When the feedback provider is tailored to the task domain, the base model is better able to extract actionable guidance, leading to enhanced performance.

\subsection{Error Analysis}
We randomly sample 50 GPT-4o outputs from our development set under each of the three settings (post-hoc evaluation, end-to-end generation, and iterative refinement), giving 150 outputs in total. From these, we identify five main categories of errors. 
Figure \ref{fig:error_analysis} provides a representative example for each error type. 

\noindent\textbf{Evidence attribution errors [25\%].} This category includes cases where the model provides redundant or invalid evidence or omits essential supporting details. Challenges in accurately identifying and extracting relevant evidence from  paragraphs lead to low precision and recall, hindering the reliability of responses.

\noindent\textbf{Execution errors [22\%].} These errors occur when the model generates non-executable code due to syntax or logical issues. 
To better understand the causes of code execution failures, we conducted a detailed human analysis of 50 randomly selected failure cases. This analysis revealed three primary error categories, as shown in Figure \ref{fig:code_error}.

\noindent\textbf{Numerical information extraction and calculation errors [20\%].} These errors involve the incorrect extraction of key financial terms, the use of mismatched units (e.g., when units are provided in table headers), and calculation mistakes such as rounding errors. 

\noindent\textbf{Knowledge validation errors [15\%].} Errors arise when the model leverages external knowledge without proper citation. This leads to challenges in maintaining accuracy and consistency in validating facts.

\noindent\textbf{Fluency, factual consistency and reasoning Errors [12\%].} Errors in this category include generating incorrect timestamps or dates, confusing monetary units, mixing data from different companies, offering faulty reasoning that lacks logical support, and outright hallucinations. 

\noindent\textbf{Others [6\%].} Other errors include overly lengthy or empty answers and instances where responses are produced in languages other than English. 


\section{Conclusion}
This paper introduces \ours, a comprehensive benchmark designed to evaluate the ability of LLMs to generate factually grounded and well-attributed answers in long-form question answering. We design fine-grained automatic evaluation methods that measure answer quality at both token and semantic levels, assess numerical correctness, and evaluate attribution quality across evidence, code, and professional knowledge. Our extensive experiments compare different generation strategies, including post-hoc, end-to-end generation and iterative refinement. Experiment results show that GPT-4o achieves the highest overall performance, while open-source models demonstrate strong potential. Additionally, \ours underscores the importance of fine-grained metrics in financial long-form question answering task. Experiments on \ours indicate that post-hoc and end-to-end generation perform similarly, as frontier LLMs can handle complex attribution tasks effectively. Furthermore, iterative self-feedback does not significantly improve overall performance, except in execution-based tasks where explicit feedback from generated outputs aids refinement. Our analysis also shows that when external feedback is introduced, models with sufficient capacity can leverage it effectively, and feedback from domain-specific financial models further improves outcomes.

\section*{Limitations}
Our primary objective is to assess models' ability to generate attributable financial report analysis. Given the novelty of this task, we begin with a simplified setting involving only two companies. While this controlled setup facilitates initial evaluation, our results indicate that current LLM-based systems still struggle with key challenges, particularly numerical reasoning and factual grounding.
Expanding the dataset to include more companies is a promising direction for future work, and we plan to pursue this as models advance and show improved performance in the current setting.

\section*{Acknowledgements}

Tiansheng Hu and Chen Zhao were supported by NYU Shanghai Center for Data Science. This work was supported in part through the NYU IT High Performance Computing resources, services, and staff expertise.

\bibliography{custom}

\appendix
\clearpage
\onecolumn
\section{Appendix}
\begin{table*}[h]
\centering
\small
\resizebox{0.9\textwidth}{!}{%
\begin{tabular}{clllc}
\toprule
\textbf{Annotator ID} & \textbf{Finance Industry Experience} & \textbf{Annotation Work}\\
\midrule

1 & 1 working and 2 internship at US & Data Annotation \\

2 & 2 working and >= 2 internship at US & Data Annotation \\

3 & 1 working at UK and 2 internship at US & Data Annotation \\

4 & 1 working and >= 3 internship at US & Data Annotation \\

5 & 2 internship at US, 1 internship at Canada & Data Annotation \\

6 & Graduate student majored in finance & Data Annotation \\

7 & 1 working at Singapore &  Annotation validation \\

8 & 1 internship at US, 3 internship at China & Annotation validation \\

9 & 1 working at Canada &  Annotation validation \\

10 & Graduate student majored in data science & Code annotation \\

11 & Graduate student majored in computer science & Code annotation \\

\bottomrule
\end{tabular}
}
\caption{Details of annotators involved in dataset construction.}
\label{profile}
\end{table*}

\begin{table*}[h]
\centering
\resizebox{0.88\textwidth}{!}{%
\renewcommand{\arraystretch}{1.1}
\begin{tabular}{lllp{10cm}}
\toprule
Organization & Model & Size & Source \\
\midrule
\multirow{1}{*}{OpenAI} & GPT-4o & -- & \texttt{gpt-4o-2024-08-06} \\

\noalign{\vskip 0.5ex}\hdashline\noalign{\vskip 0.5ex}

\multirow{1}{*}{Alibaba} & Qwen2.5 & 72B &  \texttt{Qwen/Qwen2.5-72B-Instruct} \\

\noalign{\vskip 0.5ex}\hdashline\noalign{\vskip 0.5ex}

\multirow{2}{*}{Meta} & Llama-3.3 & 70B & \texttt{meta-llama/Llama-3.3-70B-Instruct} \\
 & Llama-3.2 & 1 \& 3B & \texttt{meta-llama/Llama-3.2-*B-Instruct} \\

\noalign{\vskip 0.5ex}\hdashline\noalign{\vskip 0.5ex}
 
\multirow{2}{*}{Mistral AI} & Mistral-Small & 24B & \texttt{mistralai/Mistral-Small-24B-Instruct-2501} \\
 & Mistral & 8x22B & \texttt{mistralai/Mixtral-8x22B-Instruct-v0.1} \\

\noalign{\vskip 0.5ex}\hdashline\noalign{\vskip 0.5ex}

\multirow{1}{*}{Microsoft} & phi-4 & 14B & \texttt{microsoft/phi-4} \\

\bottomrule
\end{tabular}
}
\caption{Details of the LLMs evaluated in this study. }
\label{tab:model_specification}
\end{table*}

\begin{table*}[h!]
\centering
\resizebox{1.02\textwidth}{!}{%
\renewcommand{\arraystretch}{1.1}
\begin{tabular}{llccccccc}
\toprule
\textbf{Model} & \textbf{Setting} & \textbf{Evidence F1} & \textbf{\% Exec. Success} & \textbf{Knowl. Recall} & \textbf{R-L} & \textbf{BS} & \textbf{LLM-as-a-judge} & \textbf{Numerical F1} \\
\midrule
Llama-3.1-8B & end-to-end & 35.1 & 14.1 & 10.8 & 20.9 & 85.7 & 9.8 & 27.0 \\
Llama-3.1-8B & Iterative w/ self-feedback & 32.6 & 22.0 & 9.4 & 21.5 & 86.0 & 9.2 & 25.6 \\
Llama-3.1-8B & Iterative w/ Fino1-8B feedback & 37.2 & 24.2 & 11.2 & 22.8 & 87.4 & 10.0 & 30.9 \\
\midrule
Llama-3.2-1B & end-to-end & 1.9 & 1.5 & 0.5 & 15.3 & 84.6 & 5.6 & 11.7 \\
Llama-3.2-1B & Iterative w/ self-feedback & 0.6 & 8.5 & 0.0 & 13.9 & 81.1 & 4.5 & 6.6 \\
Llama-3.2-1B & Iterative w/ Llama-3.3-70B feedback & 1.9 & 8.5 & 0.5 & 14.3 & 80.3 & 4.4 & 5.1 \\
\midrule
Llama-3.2-3B & end-to-end & 15.1 & 22.5 & 6.2 & 18.5 & 85.8 & 7.5 & 19.9 \\
Llama-3.2-3B & Iterative w/ self-feedback & 10.0 & 22.5 & 5.0 & 20.6 & 86.8 & 6.7 & 18.5 \\
Llama-3.2-3B & Iterative w/ Llama-3.3-70B feedback & 17.0 & 19.9 & 6.0 & 22.4 & 86.8 & 8.0 & 22.9 \\
\bottomrule
\end{tabular}
}
\caption{Iterative refinement in different settings.}
\label{tab:iterative_refinement}
\end{table*}

\begin{table*}[!t]
\centering
\resizebox{0.95\textwidth}{!}{%
\addtolength{\tabcolsep}{0.1em}
\begin{tabular}{lccccccccccc}
\toprule
 \multirow{3}{*}{\textbf{Model}} & \multicolumn{5}{c}{\textbf{Attribution Performance}} & \multicolumn{6}{c}{\textbf{Answer Performance}} \\
\cmidrule(lr){2-6} \cmidrule(lr){7-12} 
 & \multicolumn{3}{c}{\textbf{Evidence}} & \textbf{\% Exec.} & \textbf{Knowl.} & \multirow{2}{*}{\textbf{R-L}} & \multirow{2}{*}{\textbf{BS}} & \textbf{LLM-} & \multicolumn{3}{c}{\textbf{Numerical}}\\
\cmidrule(lr){2-4} \cmidrule(lr){10-12} 
& \textbf{Prec.} & \textbf{Recall} & \textbf{F1} & \textbf{Success} & \textbf{Recall} & & & \textbf{as-a-judge} & \textbf{Prec.} & \textbf{Recall} & \textbf{F1}\\
\midrule
\multicolumn{12}{c}{\emph{\textbf{Post-hoc}}} \\\noalign{\vskip 0.5ex}
GPT-4o & 51.0 & \textbf{75.8} & 56.5 & 9.4 & 18.1 & 23.9 & 87.7 & 1\textbf{3.5} & 38.5 & \textbf{59.8} & \textbf{43.4} \\
Qwen2.5-72B & 40.9 & 69.0 & 46.6 & 8.8 & 18.4 & 22.9 & 88.0 & 12.9 & 35.4 & 57.4 & 40.1 \\
Llama-3.3-70B & 47.2 & 72.5 & 52.9 & 19.0 & 16.9 & 20.3 & 87.7 & 12.9 & 36.5 & 56.8 & 41.1 \\
Llama-3.2-1B & 1.4 & 2.3 & 1.5 & 1.6 & 1.0 & 16.9 & 85.1 & 5.5 & 13.7 & 15.9 & 13.4 \\
Llama-3.2-3B & 10.5 & 18.1 & 11.8 & 12.9 & 8.4 & 19.6 & 86.6 & 8.8 & 25.5 & 35.4 & 26.9 \\
Mistral-Small-24B & 52.1 & 69.2 & 55.3 & 5.2 & 18.9 & 25.8 & 88.4 & 12.5 & 38.4 & 56.0 & 41.9 \\
Mistral-8x22B & 38.5 & 60.4 & 42.9 & 9.5 & 19.3 & 22.3 & 87.5 & 12.0 & 36.1 & 47.3 & 37.3 \\
phi-4 & 41.6 & 63.7 & 46.1 & 10.0 & 17.8 & 20.9 & 87.2 & 12.8 & 34.8 & 54.1 & 39.3 \\
\midrule
\multicolumn{12}{c}{\emph{\textbf{End-to-end Generation}}} \\\noalign{\vskip 0.5ex}
GPT-4o & 49.6 & 74.6 & 55.2 & 26.9 & 16.4 & 20.2 & 86.4 & 13.3 & 37.6 & 57.8 & 41.8 \\
Qwen2.5-72B & 48.6 & 68.8 & 50.1 & 15.4 & 17.2 & 21.9 & 87.0 & 11.9 & 33.3 & 45.5 & 36.1 \\
Llama-3.3-70B & 55.0 & 69.7 & 57.9 & 20.5 & 15.5 & 21.4 & 87.2 & 11.9 & 37.7 & 49.3 & 39.6 \\
Llama-3.2-1B & 1.3 & 1.5 & 1.3 & 10.4 & 0.6 & 14.2 & 81.5 & 4.3 & 5.0 & 7.7 & 5.3 \\
Llama-3.2-3B & 11.7 & 19.6 & 13.2 & 21.3 & 8.5 & 18.9 & 85.3 & 7.4 & 19.3 & 27.8 & 20.8 \\
Mistral-Small-24B & 57.1 & 70.0 & \textbf{59.2} & 13.9 & \textbf{21.1} & 24.0 & 87.9 & 12.2 & 39.0 & 52.9 & 40.8 \\
Mistral-8x22B & 52.6 & 69.1 & 55.8 & 12.9 & 20.4 & 23.5 & 87.5 & 12.4 & 33.6 & 47.5 & 36.5 \\
phi-4 & 44.1 & 60.7 & 47.4 & 18.2 & 17.4 & 22.0 & 86.8 & 12.4 & 38.1 & 48.7 & 39.4 \\
\midrule
\multicolumn{12}{c}{\emph{\textbf{Iterative Refinement}}} \\\noalign{\vskip 0.5ex}
GPT-4o & 50.2 & 74.9 & 55.7 & \textbf{28.9} & 16.0 & 21.0 & 85.7 & 13.2 & 37.2 & 57.7 & 41.7 \\
Qwen2.5-72B & 46.8 & 67.9 & 52.1 & 22.5 & 17.4 & 22.4 & 86.6 & 12.1 & 36.1 & 46.2 & 36.6 \\
Llama-3.3-70B & 52.9 & 69.3 & 56.3 & 28.3 & 15.8 & 22.3 & 86.7 & 12.1 & 36.6 & 49.7 & 39.0 \\
Llama-3.2-1B & 0.6 & 0.6 & 0.5 & 9.6 & 0.0 & 13.5 & 81.2 & 4.1 & 3.7 & 4.6 & 3.6 \\
Llama-3.2-3B & 10.1 & 14.0 & 10.7 & 20.3 & 6.5 & 20.6 & 85.7 & 6.5 & 16.7 & 20.0 & 16.6 \\
Mistral-Small-24B & \textbf{56.4} & 68.4 & 58.2 & 20.0 & 19.0 & 25.8 & 87.7 & 12.0 & \textbf{40.1} & 51.2 & 41.2 \\
Mistral-8x22B & 43.2 & 57.6 & 46.4 & 17.9 & 16.2 & 25.3 & 87.9 & 12.4 & 34.1 & 46.9 & 37.2 \\
phi-4 & 45.4 & 60.3 & 48.4 & 18.2 & 17.7 & 25.4 & 87.5 & 12.3 & 39.1 & 49.8 & 40.4 \\ 
\bottomrule
\end{tabular}
}
\caption{Results of \emph{test} set. R-L denotes ROUGE-L, BS denotes BERTScore. The LLM-as-a-judge evaluation uses a 15-point scale, consisting of three main criteria: accuracy, numerical correctness, and supporting evidence. Each criterion is scored from 1 to 5 points. The detailed breakdown of results for each criterion is shown in the Appendix Table \ref{llm_test}.}
\label{test_result}
\end{table*}

\begin{table*}[!t]
\centering
\resizebox{0.80\textwidth}{!}{%
\addtolength{\tabcolsep}{0.1em}
\begin{tabular}{lccc}
\toprule
 \multirow{2}{*}{\textbf{Model}} & \multicolumn{3}{c}{\textbf{LLM-as-a-judge}}\\
\cmidrule(lr){2-4} 
 
 & Accuracy & Numerical Correctness & Evidence Entailment \\

\midrule
\multicolumn{4}{c}{\emph{\textbf{Post-hoc}}} \\\noalign{\vskip 0.5ex}
GPT-4o & 4.6 & 4.4 & 4.6 \\
Qwen2.5-72B & 4.4 & 4.2 & 4.4 \\
Llama-3.3-70B & 4.4 & 4.3 & 4.4 \\
Llama-3.2-1B & 1.9 & 1.8 & 1.9 \\
Llama-3.2-3B & 2.9 & 3.0 & 3.0 \\
Mistral-Small-24B & 4.2 & 4.1 & 4.2 \\
Mistral-8x22B & 4.0 & 4.1 & 4.0 \\
phi-4 & 4.4 & 4.2 & 4.4 \\

\midrule
\multicolumn{4}{c}{\emph{\textbf{End-to-end Generation}}} \\\noalign{\vskip 0.5ex}
GPT-4o & 4.7 & 4.4 & 4.6 \\
Qwen2.5-72B & 4.2 & 4.1 & 3.8\\
Llama-3.3-70B & 4.1 & 4.0 & 3.9 \\
Llama-3.2-1B & 1.5 & 1.5 & 1.5 \\
Llama-3.2-3B & 2.4 & 2.6 & 2.5 \\
Mistral-Small-24B & 4.2 & 4.1 & 4.2 \\
Mistral-8x22B & 4.0 & 3.9 & 4.1 \\
phi-4 & 4.2 & 4.1 & 4.2 \\

\midrule
\multicolumn{4}{c}{\emph{\textbf{Iterative Refinement}}} \\\noalign{\vskip 0.5ex}
GPT-4o & 4.6 & 4.3 & 4.5 \\
Qwen2.5-72B & 4.1 & 3.9 & 4.1\\
Llama-3.3-70B & 4.1 & 4.0 & 4.0 \\
Llama-3.2-1B & 1.3 & 1.3 & 1.3 \\
Llama-3.2-3B & 2.2 & 2.3 & 2.3 \\
Mistral-Small-24B & 4.1 & 4.0 & 4.1 \\
Mistral-8x22B & 4.0 & 4.1 & 4.0\\
phi-4 & 4.2 & 4.0 & 4.2 \\ 


\bottomrule
\end{tabular}
}
\caption{Breakdown results of LLM-as-a-judge at \emph{development} set.}
\label{llm_development}
\end{table*}

\begin{table*}[!t]
\centering
\resizebox{0.80\textwidth}{!}{%
\addtolength{\tabcolsep}{0.1em}
\begin{tabular}{lccc}
\toprule
 \multirow{2}{*}{\textbf{Model}} & \multicolumn{3}{c}{\textbf{LLM-as-a-judge}}\\
\cmidrule(lr){2-4} 
 
 & Accuracy & Numerical Correctness & Evidence Entailment \\

\midrule
\multicolumn{4}{c}{\emph{\textbf{Post-hoc}}} \\\noalign{\vskip 0.5ex}
GPT-4o & 4.6 & 4.3 & 4.5 \\
Qwen2.5-72B & 4.4 & 4.1 & 4.4 \\
Llama-3.3-70B & 4.4 & 4.2 & 4.4 \\
Llama-3.2-1B & 1.9 & 1.7 & 1.9 \\
Llama-3.2-3B & 2.9 & 2.9 & 3.0 \\
Mistral-Small-24B & 4.2 & 4.1 & 4.2 \\
Mistral-8x22B & 4.0 & 4.0 & 4.0 \\
phi-4 & 4.3 & 4.1 & 4.4 \\

\midrule
\multicolumn{4}{c}{\emph{\textbf{End-to-end Generation}}} \\\noalign{\vskip 0.5ex}
GPT-4o & 4.5 & 4.3 & 4.5 \\
Qwen2.5-72B & 3.9 &  3.9 & 4.1 \\
Llama-3.3-70B & 4.0 & 3.9 & 3.9 \\
Llama-3.2-1B & 1.5 & 1.4 & 1.4 \\
Llama-3.2-3B & 2.4 & 2.4 & 2.5 \\
Mistral-Small-24B & 4.1 & 4.0 & 4.1 \\
Mistral-8x22B & 4.2 & 4.0 & 4.2\\
phi-4 & 4.2 & 4.0 & 4.2 \\

\midrule
\multicolumn{4}{c}{\emph{\textbf{Iterative Refinement}}} \\\noalign{\vskip 0.5ex}
GPT-4o & 4.5 & 4.2 & 4.5 \\
Qwen2.5-72B & 4.1 & 4.0 & 4.1 \\
Llama-3.3-70B & 4.1 & 4.0 & 4.0 \\
Llama-3.2-1B & 1.4 & 1.4 & 1.3 \\
Llama-3.2-3B & 2.2 & 2.2 & 2.2 \\
Mistral-Small-24B & 4.0 & 3.9 & 4.0 \\
Mistral-8x22B & 4.3 & 4.1 & 4.0 \\
phi-4 & 4.2 & 4.0 & 4.2 \\ 


\bottomrule
\end{tabular}
}
\caption{Breakdown results of LLM-as-a-judge at \emph{test} set.}
\label{llm_test}
\end{table*}

\begin{figure*}[h!]
\begin{tcolorbox}[colback=black!7.5!white, colframe=black!40!black, title=Answer Generation, fontupper=\footnotesize, fonttitle=\footnotesize]

You are a professional financial analyst. Your task is to analyze financial reports from two companies and answer a specific question based on the provided information. Your response must be well-structured, concise, and fact-based. \\
    
When presenting your analysis, structure your response into numbered clauses. \\

Each clause should be a single concise sentence presenting a fact-based financial insight. \\

Avoid sub-bullets, section headers, or formatting symbols (such as dashes, asterisks, or bold text). \\

Write in full sentences. Do not introduce bullet points or separate explanations within the same clause. \\

Here is an example of response: \emph{example} \\

The following are the financial reports of two companies. \\

Financial Report for Company \textbf{<name1>}: \\
\textbf{<report1>} \\

Financial Report for Company \textbf{<name2>}: \\
\textbf{<report2>} \\

Based on the information provided, answer the following question:
Question: \textbf{<question>}

\end{tcolorbox}

\caption{Prompt of post-hoc answer generation.}
\label{post_hoc_ans}
\end{figure*}

\begin{figure*}[h!]
\begin{tcolorbox}[colback=black!7.5!white, colframe=black!40!black, title=Attribution Generation, fontupper=\footnotesize, fonttitle=\footnotesize]

Given the financial reports of two companies, a question, an answer to the question structured into numbered clauses, and a list of professional knowledge, your task is to identify three attributes for each clause. \\
    
The three attributes are: \\
1. The evidence supporting the clause. It can be a list of paragraph indices from the company's financial report or inferred from previous clauses. \\
2. Code: If the clause involves numerical values that being calculated from the context, provide a Python function that takes necessary input values, performs the calculation, and returns the result. \\
3. Professional Knowledge: Identify the professional knowledge items that are relevant to the clause from the professional knowledge list if any. \\

Here is an example of response: \emph{example} \\

The following are the financial reports of two companies, a question, and the answer to the question in the numbered clauses format, and a list of professional knowledge. Based on the information provided, identify three attributes for each clause. \\

Financial Report for Company \textbf{<name1>}: \\
\textbf{<report1>} \\

Financial Report for Company \textbf{<name2>}: \\
\textbf{<report2>} \\

Question: \textbf{<question>} \\

Answer to the Question: \\
\textbf{<answer>} \\

Professional Knowledge List: \textbf{<knowledge list>} \\

Attributes for each clause: \\

\end{tcolorbox}
\caption{Prompt of post-hoc attribution generation.}
\label{post_hoc_attr}
\end{figure*}

\begin{figure*}[h!]
\begin{tcolorbox}[colback=black!7.5!white, colframe=black!40!black, title=End-to-end Generation, fontupper=\footnotesize, fonttitle=\footnotesize]

You are a professional financial analyst. Your task is to analyze financial reports from two companies to answer a specific question based on the provided information. For each clause in your response, provide the following three attributes: \\
1. Evidence: Specify supporting evidence as paragraph indices from the financial reports or indicate if it’s inferred from previous clauses. \\
2. Code: If numerical values are calculated, provide a Python function that performs the calculation and returns the result.\\
3. Professional Knowledge: Identify relevant professional knowledge items from the provided list, if applicable.\\

Structure your response into numbered clauses, each with its corresponding attributes.\\
Here is an example of response: \emph{example} \\

The following are the financial reports of two companies, a question, and a list of professional knowledge. Based on the information provided, please answer the question in the format of numbered clauses, each with three attributes: Evidence, Code, and Professional Knowledge. \\

Financial Report for Company \textbf{<name1>}: \\
\textbf{<report1>} \\

Financial Report for Company \textbf{<name2>}: \\
\textbf{<report2>} \\

Question: \textbf{<question>} \\

Professional Knowledge List: \textbf{<knowledge list>}\\

Your response:

\end{tcolorbox}

\caption{Prompt of end-to-end generation.}
\label{e2e_generation}
\end{figure*}

\begin{figure*}[h!]
\begin{tcolorbox}[colback=black!7.5!white, colframe=black!40!black, title=Iterative Refinement, fontupper=\footnotesize, fonttitle=\footnotesize]

You are a professional financial analyst responsible for reviewing and improving a financial analysis. You will be given: \\

1. Financial reports from two companies.\\
2. A question related to their financial performance.\\
3. A current analysis, structured as numbered clauses. Each clause consists of:\\
    - Clause Content: A clear and concise financial insight.\\
    - Attributes:\\
        1) Evidence: References to supporting paragraph indices from the financial reports or logical inference from previous clauses.\\
        2) Code: A Python function performing necessary calculations, including an execution result for verification.\\
        3) Professional Knowledge: Relevant financial concepts from the provided knowledge list.\\

Provide constructive feedback by evaluating each clause based on the following criteria:\\
1. Conciseness \& Relevance to the Question\\
- Does the analysis answer the question directly and efficiently?\\
- Does it include unnecessary details or over-explained information? If so, suggest more concise phrasing.\\
2. Numerical Accuracy\\
- Verify whether numerical values are correctly calculated based on the financial reports.\\
- Ensure that all provided calculations are necessary and relevant to answering the question.\\
3. Evidence Support \& Justification\\
- Does each clause rely on valid evidence from the reports? If inference is used, is it logically sound?\\
- Are financial concepts correctly applied according to the provided knowledge list?\\
- Does the analysis remain within the provided information, avoiding speculation or unsupported conclusions?\\
4. Code Accuracy: \\
- Is the provided code correct, and does its execution result align with the content?\\
- Professional Knowledge Validity: Are the cited financial concepts appropriate and correctly applied?\\
Clearly state specific issues and provide actionable suggestions for improvement.\\
If no corrections are needed, output "Done" without any additional text.\\

The following are the financial reports of two companies, a question, a list of professional knowledge, and a financial analysis. Based on the information provided, please provide a feedback. If no additional feedback is necessary, output "Done" without any other text. \\

Financial Report for Company \textbf{<name1>}: \\
\textbf{<report1>} \\

Financial Report for Company \textbf{<name2>}: \\
\textbf{<report2>} \\

Question: \textbf{<question>} \\

Professional Knowledge List: \textbf{<knowledge list>}\\

Analysis: \textbf{<analysis>} \\

Your response:

\end{tcolorbox}

\caption{Prompt of iterative refinement.}
\label{e2e_feedback}
\end{figure*}

\clearpage
\begin{figure*}[h!]
\centering
\begin{tcolorbox}[
  colback=black!7.5!white,
  colframe=black!40!black,
  title={Code Error Analysis},
  fontupper=\footnotesize,
  fonttitle=\footnotesize,
]
\renewcommand{\arraystretch}{1.2}
\begin{tabularx}{\textwidth}{%
  >{\footnotesize\raggedright\arraybackslash}p{2.5cm}%
  >{\footnotesize\raggedright\arraybackslash}X%
  >{\footnotesize\raggedright\arraybackslash}p{2.5cm}%
}
\toprule
\textbf{Error Category} & \textbf{Example} & \textbf{Explanation} \\
\midrule
Calling Functions Without Providing Required Arguments (46\%) &
\begin{tabular}[t]{@{}>{\footnotesize\raggedright\arraybackslash}X@{}}
\textbf{Code:}\\[1mm]
\texttt{def net\_interest\_income\_sensitivity(
base\_year\_net\_interest\_income, 
change\_in\_basis\_points):}\\
\texttt{\ \ \ \ if change\_in\_basis\_points == 200:}\\
\texttt{\ \ \ \ \ \ \ \ sensitivity = -3.1}\\
\texttt{\ \ \ \ else:}\\
\texttt{\ \ \ \ \ \ \ \ sensitivity = None}\\
\texttt{\ \ \ \ return sensitivity}
\end{tabular}
&
These errors occur when functions are defined with parameters but are called without supplying the necessary inputs. \\[3mm]
\midrule
Missing Return Statements After Performing Calculations (20\%) &
\begin{tabular}[t]{@{}>{\footnotesize\raggedright\arraybackslash}X@{}}
\textbf{Code:}\\[1mm]
\texttt{def compare\_depreciation\_increase():}\\
\texttt{\ \ \ \ cwt\_increase = 2.9}\\
\texttt{\ \ \ \ awk\_increase = 15}
\end{tabular}
&
In these cases, the function performs some internal computation but does not return the result, leading to \texttt{None} outputs. \\[3mm]
\midrule
Indentation Errors (16\%) &
\begin{tabular}[t]{@{}>{\footnotesize\raggedright\arraybackslash}X@{}}
\textbf{Code:}\\[1mm]
\texttt{def debt\_to\_equity\_ratio\_CWT():}\\
\texttt{\ \ \ \ \ \ total\_debt = 1051952 + 413610}\\
\texttt{\ \ \ \ \ \ total\_equity = 4357905 - 3129132}\\
\texttt{\ \ \ \ \ \ return total\_debt / total\_equity}
\end{tabular}
&
These errors are likely caused by the model generating long outputs where text and code are interleaved, leading to incorrect indentation. \\[3mm]
\bottomrule
\end{tabularx}
\end{tcolorbox}
\caption{Code Error Analysis}
\label{fig:code_error}
\end{figure*}

\clearpage 
\begin{figure*}[h!]
\centering
\begin{tcolorbox}[
  colback=black!7.5!white,
  colframe=black!40!black,
  title={Error Analysis Table},
  fontupper=\footnotesize,
  fonttitle=\footnotesize,
]
\renewcommand{\arraystretch}{0.9} 
\begin{tabularx}{\textwidth}{%
  >{\footnotesize\raggedright\arraybackslash}p{3cm}%
  >{\footnotesize\raggedright\arraybackslash}X%
  >{\footnotesize\raggedright\arraybackslash}p{4cm}%
}
\toprule
\textbf{Error Category} & \textbf{Example} & \textbf{Explanation} \\
\midrule
Evidence Attribution Errors &
\begin{tabular}[t]{@{}p{6cm}@{}}
\textbf{Model Answer:}\\
… As of March 31, 2024, CWT's issuer had total assets of \$4.36 billion, while AWK's specific figures are not provided in detail. …\\[1mm]
{\small\textbf{Evidence:}} {\small 'CWT': [15], 'AWK': [4, 5, 10, 11]}\\[1mm]
{\small\textbf{Ground Truth:}} {\small 'CWT': [14, 15], 'AWK': [3, 4]}\\[1mm]
\end{tabular}
&
In the ground truth, it cites the 14th and 15th fragments from CWT as well as the 4th and 5th fragments from AWK. However, in the model answer, it omits citing the 14th fragment from CWT and redundantly cites the 10th and 11th fragments from AWK. \\[3mm]
\midrule
Execution Errors &
\begin{tabular}[t]{@{}p{6cm}@{}}
\textbf{Model Answer Code:}\\[1mm]
\texttt{def compute\_percentage(part, total):}\\
\texttt{\ \ \ \ return (part / total) * 100}\\[1mm]
\end{tabular}
&
Functions are defined with parameters but are called without supplying the necessary inputs. \\[3mm]
\midrule
Numerical Information Extraction and Calculation Errors &
\begin{tabular}[t]{@{}p{6cm}@{}}
\textbf{Model Answer:}\\
… CWT's issuer had net utility plant assets comprising approximately 82\% of total assets (\$3.55 billion out of \$4.36 billion).\\[1mm]
\textbf{Execution Result:}\\
81.4\\
\end{tabular}
&
The rounded value of model is incorrect, leading to a calculation error. \\[3mm]
\midrule
Knowledge Validation Errors &
\begin{tabular}[t]{@{}p{6cm}@{}}
\textbf{Model Answer:}\\
… CVX reported a net income of \$5.5 billion in Q1 2024, down 16.3\% from \$6.6 billion in Q1 2023.… \\[1mm]
\textbf{Professional Knowledge:}\\
``Net Profit Margin = Net Income / Revenue''
\end{tabular}
&
Irrelevant professional knowledge was cited. In the clause, it only states the absolute net income figures and their year-over-year change. However, the cited professional knowledge is about the Net Profit Margin. This is a wrong citation, resulting in a knowledge validation error. \\[3mm]

\midrule
Fluency, Factual Consistency and Reasoning Errors &
\begin{tabular}[t]{@{}p{6cm}@{}}
\textbf{Model Answer:}\\
… ZS's ongoing losses highlight its growth-focused strategy, while CVX maintains profitability, supported by its strong market presence.\\[1mm]
\textbf{Evidence:}\\
{'ZS': [6], 'CVX': [6, 12]}
\end{tabular}
&
This sentence is a predictive statement derived from inferences made on preceding clauses, so its attribution should be considered inference rather than evidence-based. This attribution error reveals that the model failed to accurately identify the function of the sentence, indicating an error in its reasoning. \\[3mm]
\midrule
Others &
\begin{tabular}[t]{@{}p{6cm}@{}}
\textbf{Model Answer:}\\
The Current Ratio is a measure of liquidity calculated as Current Assets divided by Current Liabilities. For Company EOG, as of March 31, 2024, Current Assets totaled \$9,928 million and Current Liabilities totaled \$4,273 million. EOG's Current Ratio is approximately 2.32 (\$9,928 million / \$4,273 million). For Company NOG, as of March 31, 2024, Current Assets totaled \$412,796 thousand and Current Liabilities totaled \$428,010 thousand. NOG's Current Ratio is approximately 0.96 (\$412,796 thousand / \$428,010 thousand). A Current Ratio above 1 indicates that a company is better positioned to cover its short-term obligations, while a ratio below 1 indicates potential liquidity challenges.\\
\textbf{Ground Truth:}\\
EOG's Current Ratio is 2.32 and NOG's is 0.96.
\end{tabular}
&
The answer is overly verbose and redundant compared with the concise ground truth.
 \\

\bottomrule
\end{tabularx}
\end{tcolorbox}
\caption{Error Analysis Table}
\label{fig:error_analysis}
\end{figure*}
\clearpage 

\clearpage 
\clearpage 
\end{document}